\documentclass[letterpaper, 10 pt, conference]{ieeeconf}  % Comment this line out if you need a4paper

\IEEEoverridecommandlockouts                              % This command is only needed if 
\usepackage{graphicx} % for pdf, bitmapped graphics files
\usepackage{amsmath} % assumes amsmath package installed
\usepackage{amssymb}  % assumes amsmath package installed
\usepackage[ruled,vlined]{algorithm2e}
\usepackage{amsfonts}
\usepackage{subcaption}
\usepackage{balance}
\usepackage{xcolor}
\usepackage{listings}
\usepackage{color,soul}
\definecolor{mygreen}{rgb}{0,0.6,0}
\definecolor{myorange}{rgb}{1.0,0.5,0.3}
\definecolor{mymauve}{rgb}{0.58,0,0.82}
\definecolor{myblue}{rgb}{0.05,0.19,0.57}
\definecolor{mygrey}{rgb}{0.4,0.4,0.4}
\definecolor{myred}{rgb}{0.9,0.2,0.15}

\lstdefinelanguage{pddl}
{
  sensitive=false,    % not case-sensitive
  morecomment=[l]{;}, % line comment
  alsoletter={:,-},   % consider extra characters
  morekeywords={
    define,domain,problem,not,and,or,when,forall,exists,either,
    :domain,:extends,:requirements,:types,:objects,:constants,
    :predicates,:action,:parameters,:precondition,:effect,:functions,
    :fluents,:primary-effect,:side-effect,:init,:goal,assign
    :strips,:adl,:equality,:typing,:conditional-effects, :metric, minimize,
    :negative-preconditions,:disjunctive-preconditions,
    :existential-preconditions,:universal-preconditions
  },
  keywords=[2]{object}, % Objects
  keywords=[3]{}, % Predicates
  keywords=[4]{assignMenus, assignMC, assignSC}, % Actions
  keywords=[5]{} % Functions
}

\usepackage[]{todonotes}

\newcommand{\norm}[1]{\left\lVert#1\right\rVert}

\graphicspath{{figures/}}

\title{\LARGE \bf
  Hierarchical Human-Motion Prediction and Logic-Geometric Programming
  for Minimal Interference Human-Robot Tasks
}

\author{An T. Le$^{1, *}$, Philipp Kratzer$^{1}$, Simon Hagenmayer$^{1}$, Marc Toussaint$^{2,3}$ and Jim Mainprice$^{1,2}$\\% <-this % stops a space
\vspace{0.1cm}
\authorblockA{\tt{*\small{an.thai.le97@gmail.com}}}
\authorblockA{$^1$Machine Learning and Robotics Lab, University of Stuttgart, Germany}
\authorblockA{$^2$Max Planck Institute for Intelligent Systems ;  IS-MPI ; T{\"u}bingen/Stuttgart, Germany}
\authorblockA{$^3$Technische Universit\"at Berlin ; TUB ; Germany}
\vspace{-0.8cm}
}

\begin{document}

\maketitle
\thispagestyle{empty}
\pagestyle{empty}

%%%%%%%%%%%%%%%%%%%%%%%%%%%%%%%%%%%%%%%%%%%%%%%%%%%%%%%%%%%%%%%%%%%%%%%%%%%%%%%%
\begin{abstract}

In this paper, 
we tackle the problem of human-robot coordination
in sequences of manipulation tasks.
Our approach integrates hierarchical
human motion prediction with Task and Motion Planning (TAMP).
We first devise a hierarchical motion prediction approach
by combining Inverse Reinforcement Learning and short-term motion
prediction using a Recurrent Neural Network.
In a second step, we propose a dynamic version of the TAMP
algorithm Logic-Geometric Programming (LGP) \cite{Toussaint:15}.
Our version of Dynamic LGP,
replans periodically to handle the mismatch between the human motion prediction
and the actual human behavior.
We assess the efficacy of the approach by training the prediction algorithms and testing
the framework on the publicly available MoGaze dataset \cite{kratzer2020mogaze}.

\end{abstract}

%%%%%%%%%%%%%%%%%%%%%%%%%%%%%%%%%%%%%%%%%%%%%%%%%%%%%%%%%%%%%%%%%%%%%%%%%%%%%%%%
\section{INTRODUCTION}

%Motivate
As robots become more capable,
they will increasingly share space with humans.
Consider the case where humans and robots additionally
share a task, such as caring for people in a hospital.
Robots could tidy, clean, bring lunch to patients, gather medication,
or even prepare an operation room.
The robot would be collaborating with one or several humans,
not only sharing space but also partially executing the task
the human is busy with.
In such cases the user is interested in having maximal support
from the robot while requiring a minimal amount of interference
with its own task objectives. Humans do this naturally
when they collaborate. For instance, one could put food back into the fridge while the other collects and cleans the dishes. %can tidy the kitchen while the other could take care of the living room.
In this paper we propose a framework that can produce such behaviors.
We do this by combining hierarchical human motion-prediction and
Task and Motion Planning (TAMP) framed as a Logic-Geometric Program (LGP) \cite{Toussaint:15}.

%For safe and efficient human-robot collaboration it is crucial
% to account explicitly for the human when generating the robot behavior
%\cite{Bauer:08, Kruse:13}.
%Moreover, it is important to predict the human behavior in order to coordinate the human
%and robot actions, or interact without disturbing the natural flow of the human's motion.
%This has been noted in numerous prior works \cite{ziebart2009planning, Mainprice:13}.
%%A strategy often employed is to generate robot trajectories that minimally
%%intervene with the human while still retaining the ability that both, the human and the robot,
%%can achieve their goals without having to deviate widely from their optimal/natural path.
%Hence, human intent and motion prediction is becoming an increasingly
%important topic of robotics research, which has been recently
%surveyed by Rudenko et al.~\cite{rudenko2019human}.

\begin{figure}
\centering
\includegraphics[width=\linewidth]{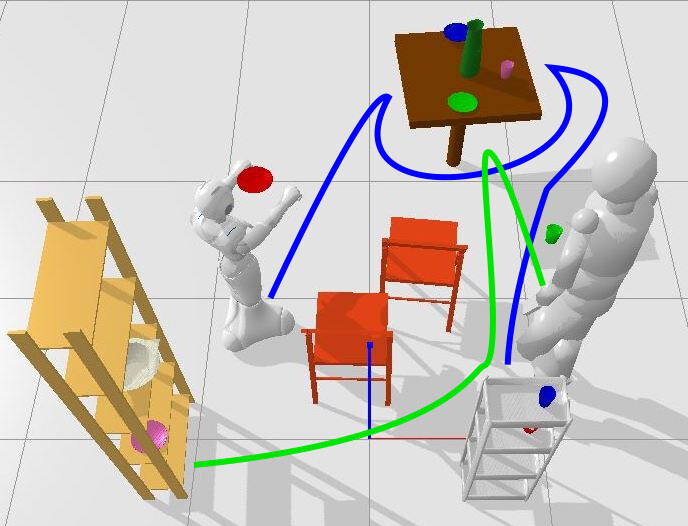}
\caption{Pepper and a human jointly setup the table. 
%Pepper carries a plate while the human is carrying a green cup.
Motion plans resulting from Dynamic LGP, minimally interfere with the human task
while reducing the overall completion time. }
\label{fig:set_table}
\end{figure}

For safe and efficient human-robot collaboration it is crucial
to account explicitly for the human when generating the robot behavior
\cite{Bauer:08, Kruse:13}.
Moreover, it is important to predict the human behavior in order to coordinate the human
and robot actions, or interact without disturbing the natural flow of the human's motion.
This has been noted in numerous prior works \cite{ziebart2009planning, Mainprice:13}.
Hence, human intent and motion prediction is becoming an increasingly
important topic of robotics research, which has been recently
surveyed by Rudenko et al.~\cite{rudenko2019human}.

A key challenge for human-motion prediction
is to predict movement over a long horizon.
This problem arises naturally in the case of sequences of manipulation
motions, for example, when a human and robot have to prepare
a table for dinner or tidy a room (see Figure \ref{fig:set_table}).
In this work, we propose a hierarchical prediction approach,
which can handle such long-term horizon.
The general approach is to use two hierarchical levels:
symbolic, (i.e. discrete) and geometric (i.e., continuous).
For the robot motion planning, we also use these two levels
in a TAMP algorithm.

As human motion is the result of complex biomechanical processes that
are challenging to model, state-of-the-art work on motion prediction focuses
on data-driven models, such as recurrent neural networks~\cite{wang2019vred, martinez2017human, pavllo2019modeling}.
Our prediction approach makes use of Maximum Entropy Inverse Reinforcement Learning \cite{ziebart2008maximum},
to produce a discrete policy, this policy decides what actions to take
next based on the symbolic states of the world.
At the lowest level we predict movements using a Recurrent Neural
Network (RNN), which is trained on the MoGaze dataset \cite{kratzer2020mogaze}.
In order to adapt the discrete policy to continuous motion prediction,
we introduce goal conditioning to the RNN VRED architecture \cite{wang2019vred}.
We combine two networks, one conditioned on hand target goals for manipulation
and another one on pelvis target goals for walking. 
The high-level policy sequences goals by means of an intermediate grounding of symbolic layer.
resulting in a single trajectory prediction.

%A drawback of these architectures is that the network is only trained on the human state and therefore not able to take scene context, such as targets for reaching motion or obstacles into account. Adapting scene context directly into the architecture would require a generalizable scene representation and huge amounts of training data to be able to generalize to unseen environments.

%Nonetheless, forecasting human whole-body motion as well as inferring
%3D human pose has received significant attention from the computer vision community,
%hence there is a large amount of available full-body motion datasets.
%A widely used dataset is the CMU Graphics Lab motion capture database,
%which contains a large amount of motion data from different humans~\cite{mocapcmu}.
%Another commonly used dataset is the Human3.6m dataset
%recorded by  Ionescu et al.~\cite{ionescu2013human3}.
%It is a large scale dataset with 11 individual actors that is
%often used to evaluate short-term motion prediction algorithms.
%A large collection of short motion trajectories is the KIT Whole-Body Human Motion Database~\cite{mandery2015kit}. It can be used with a unified representation of human motion and motions can be transferred to humanoid robots~\cite{mandery2016unifying}.  The database contains motions including objects, such as drinking, shaking, pouring, and throwing of objects.

Given this prediction, the planning module makes use of
LGP to produce minimally interfering plans which support the
human with the task objectives.
There we devise a symbolic representation 
of the workspace and the task goal using the Planning Domain Definition Language (PDDL)~\cite{McDermott:98}. Given both the geometric and the symbolic
representation, LGP explores the space of skeletons using tree search,
which results in a heuristic ranking of symbolic plans.
Skeletons are then evaluated
with increasing level of accuracy until a feasible plan is found.
At the lowest-level a full trajectory is planned using non-linear
programming.
The optimizer uses an interior point method and a ``Gauss-Newton" approximation of the Hessian \cite{Mainprice:20interior, Mainprice:21bewego}.

We combine human motion prediction and robot motion planning by predicting changes in the symbolic state and planning in the combined symbolic state. We first predict what the human would do and
plan for the actions that the human would do last.
In order to handle erroneous human motion prediction, we extend the basic
formulation of LGP to handle dynamic changes in the workspace.

To summarize the main contributions of the paper:
\begin{itemize}
\item We propose a new formulation to produce long-term 
task sequences for a human-robot team that support the human while minimizing the
interference.
\item We introduce a new hierarchical motion prediction system
which is able to produce full-body prediction in long horizons
\item We present results assessing the efficacy of our approach
using the MoGaze \cite{kratzer2020mogaze} dataset.
\end{itemize}

This paper is organized as follows: In Section~\ref{sec:related_work}
we discuss relevant prior work.
Section~\ref{sec:method} introduces our framework,
and gives information concerning our implementation.
In Section~\ref{sec:experiments} we evaluate our framework 
on motion capture data.
Conclusions are drawn in Section~\ref{sec:conclusions}.

\section{Related Work}
\label{sec:related_work}

\subsection{Human-Robot Collaboration}
Human-Robot Collaboration (HRC) focuses on robotic systems able
to perform joint actions with humans \cite{Bauer:2008bx, Ajoudani:2017kl}. The robot is a member of a mixed human-robot team, where members share a common goal.
In this context, shared task planning and interactive motion planning allow for higher
level collaborations and is thus a topic of interest in HRC.

In order to schedule coordinated actions,
a lot of work has explored how to model the capabilities of the agents
in the workspace \cite{Hoffman:07}.
For intelligently account for space-sharing
conventions the notion of \textit{Proxemics} is now well accepted in HRC~\cite{Kruse:2013jx}.

Some works include high-level symbolic planning in order to find a human-aware robot plan \cite{alami2005task, lemaignan2017artificial} and also combining task and motion planning has shown success for human-aware HRC \cite{Gharbi:2015iz, Busch:2018ix}.
However, no work proposes to integrate a full hierarchical predictive model of human behavior.

\subsection{Human-Motion Prediction}
Human movement is the
result of simultaneous control of
hundreds of degrees of freedom. However, muscles
are controlled in coordination (i.e., synergies),
which yields a low dimensional embedding of motion.
Thus, time series techniques such as HMMs, though limited to low dimensional state spaces,
have had some success in precting movement and activites \cite{kulic2012incremental}.

A lot of work in the area of movement prediction
and regression of motion capture data focuses on data driven methods.
In these applications less attention is put to reproducing physically correct forces, torques or muscle activation,
but rather reproduce plausible movement.
Nonlinear function approximators such as 
Gaussian Processes~\cite{kratzer2018towards} or Deep Neural Networks \cite{martinez2017human, fragkiadaki2015recurrent}
have been used to regress large databases of human movement. Recurrent Neural Networks (RNN) are state of the 
art for predicting short-term high dimensional movements \cite{wang2019vred, martinez2017human, kratzer2019prediction}.

However, these methods mostly consider very short motions, often less than a second, and therefore are not suitable for predicting long sequences of manipulation tasks. To address this issue, we adapt a RNN architecture and make it goal-conditioned. We use a discrete, high-level policy in order to predict intermediate goals allowing us to predict long motion sequences by chaining the short-term predictions.

\subsection{Task and Motion Planning}
In classical AI, High Level Planning  \cite{Ghallab:04} has
 been studied for decades and many languages and
planning paradigms have been developed to solve symbolic planning.

Research in Robotics has sought to integrate such concepts
for the purpose of solving motion planning problems
involving sequences of tasks such as object manipulation or footsteps.
Task and Motion Planning (TAMP) is a sub-field of motion planning and robotics that aims to find multiple intricate and sequential manipulation movements. Usually TAMP involves  reasoning  on  a  symbolic  level, which provides discrete action sequences, and continuous motion planning, which tries to find motion trajectories fulfilling the discrete action sequence.

Approaches to TAMP often are random sampling methods, \cite{Simeon:04, kaelbling2011hierarchical, dantam2018incremental},  constrained-based methods \cite{lozano2014constraint, lagriffoul2014efficiently} or numerical optimization based methods \cite{Toussaint:15}.

In this work we focus on Logic Geometric Programming (LGP) \cite{Toussaint:15}, which is an approach combining logic tree search with trajectory optimization techniques, and combine it with human motion prediction to plan a collaborative task in a scenario with a human-robot team.
The uncertainty in human motion prediction is handled by dynamic replanning, which also extends the basic formulation of LGP to handle general dynamic workspaces.
In \cite{toussaint2017multi}, a human-robot collaboration task is implemented using LGP, where
the human prediction is modeled with simple cost terms, and no replanning is performed. To our knowledge,
LGP has never been combined with a learned predictive model of human motion.

%\newpage

\section {Hierarchical Motion Prediction \& Planning}
\label{sec:method}

Here we devise the framework for TAMP
using a long-term prediction of human motion (see Figure \ref{fig:model}).
We rely on a symbolic decomposition to describe the
task, which we encode using the
Planning Domain Definition Language (PDDL) \cite{McDermott:98}.

%We first report on our extension of LGP in Section \ref{subsec:dynamiclgp}
%and then present the motion prediction framework in Section \ref{subsec:motion_prediction}.

\subsection{Dynamic LGP}
\label{subsec:dynamiclgp}
 
For motion planning, we introduce Dynamic LGP, which is a variant of LGP,
that has a replanning ability at Level 3 LGP~\cite{Toussaint:15},
which given the current environmental conditions, solves for 
the minimal interference Human-Robot tasks.

\begin{figure}[t]
  \includegraphics[width=\columnwidth]{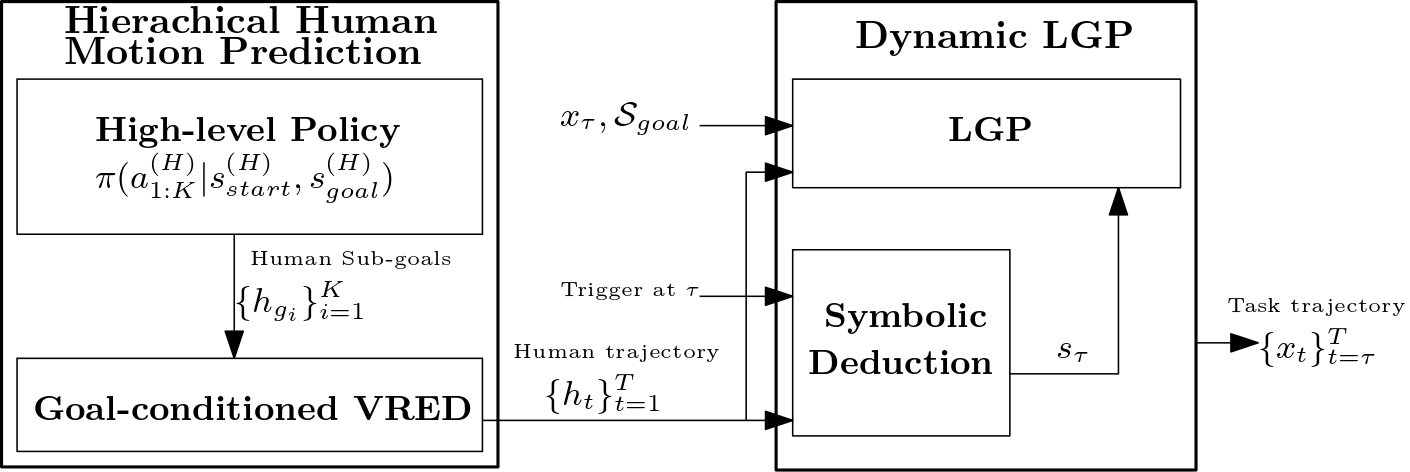}
  \caption{Dataflow through our framework. The human motion prediction 
  system produces a trajectory $\{ h_t \}_{t=1}^T$ that is being
  used to deduce the future symbolic state of world. Every replanning event is triggered at $\tau$ where a Logic-Geometric Program is optimized.}
  \label{fig:model}
%  \vspace{-.8cm}
\end{figure}

The basic idea of LGP, is to decompose the task with two
levels of abstraction. At the highest level we consider
a discrete set of actions $A =\{ a_i \}_{i=1}^N$, for instance
\textit{move}, \textit{pick} and \textit{place} (see Figure \ref{fig:domain}).
We call a \textit{skeleton}, a sequence of symbolic actions $a_{1:K}^{(R)}$.
Here we denote $(R)$ to refer to the robot sequence of actions, as it is different from the human sequence of actions $a_{1:K}^{(H)}$ used in High-level Policy (see Figure \ref{fig:model}).

A fully instantiated plan is then a skeleton, together with a
motion trajectory $x : [0, T] \to \chi$,
where $\chi = \mathcal{C} \times \mathcal{H} \times \mathcal{O}$,
the Cartesian product of the robot, human and movable object configuration spaces
respectively.

\subsubsection{Problem formulation}
\label{subsubsec:lgpprob}
  An instance $I$ of Dynamic LGP consists of the following components:

  \textbf{Symbolic Domain}:
  \begin{itemize}
      \item Predicates $\mathbb{P} = \{P_1(.), ..., P_N(.)\}$
      \item Constants $O$ as terms/arguments for predicates~$\mathbb{P}$
      \item All symbolic states $s \in \mathbb{S}$ in the domain, where each state is a set of grounded propositions from the predicates $\mathbb{P}$
      \item A set of actions $a = (R, P, E) \in A$ where:
      \begin{itemize}
          \item $R$ : parameters of the action.
          \item $P \subset \mathbb{P}$ : precondition predicates.
          \item $E \subset \mathbb{P}$ : effect predicates.
      \end{itemize}
  \end{itemize}
  In our experiments, we adopt a PDDL-syntax to describe the symbolic domain. 
  
  \textbf{Geometric problem}:
  Let $\mathcal{C}$ be the configuration space
  of the robot and the geometric state at time $t$, $x_t \in \chi$.
  The task is to find a global path $x : t \mapsto x_t$,
  which minimizes the following LGP:
  \small
  \begin{subequations}
    \begin{align} \label{eq:geoprob}
      &\underset{x, a_{1:K}^{(R)}, s_{1:K}}{min} \int_{0}^{KT} c(x(t), \dot{x}(t), \ddot{x}(t), s_{k(t)}) \text{dt} \; \\
      \text{s.t.}\nonumber \\
        &    x(0) = x_0,~h_{\text{goal}}(x(KT)) = 0,~ g_{\text{goal}}(x(KT)) \leq 0 \\
        &    \forall t \in [0, KT] : h_{p}(x(t), \dot{x}(t), s_{k(t)}) = 0, \nonumber\\
        &    \hspace{2cm}g_{p}(x(t), \dot{x}(t), s_{k(t)}) \leq 0 \\
        &    \forall k \in \{1,...,K\} : h_{sw}(x(t), \dot{x}(t), a_k) = 0 \\
        &     s_k \in \text{exec}_{a_k}(s_{k-1}) \\
        &     s_K \in \mathbb{S}_{\text{goal}}
    \end{align}
  \end{subequations}
  \normalsize
  where the path is global continuous $x$ and contains $K \in N$ phases, each has fixed duration $T > 0$.

%  The smoothness in a phase is required $x \in C^2([0, KT], \chi)$ for Jacobian and Hessian matrices computation in the solver. 
  In our experiments, the cost function $c : (q_t, \dot q_t, \ddot q_t, s) \mapsto c_t \in \mathbb{R}$,
  is a combination of differentiable maps, penalizing velocities and accelerations of the robot.
  Obstacle avoidance and goal manifold
  are enforced using equality and inequality constraints
  $h_{p},~g_{p}$ in the phase $k(t) \in [t/T]$ conditioned on a discrete symbolic state $s_k \in \mathbb{S}$.
  
  To impose transition conditions between phases, the switch functions $h_{sw},~g_{sw}$ define equalities and inequalities constraints conditioned on the transition action $a_k$.
  We assume that the equality and inequality functions are differentiable.

\subsubsection{Solving LGP}
\label{subsubsec:solvelgp}

  To search the symbolic domain for a skeleton satisfying all constraints,
  the action set $A$ has to be grounded with the constants set $O$ \cite{Fox:03},
  resulting in the grounded action set $A_g$.
  
  The most basic operations for searching are the feasibility check
  and the state transition. In this case, the operations can be formally stated as:
  \begin{itemize}
    \item \textbf{Action feasibility check}: A grounded action $a = (R, p, e) \in A_g$, in which $p, e$ are the grounded propositions of the preconditions and the effects~\cite{Fox:03}, 
    is applicable to $s$ iff $p \subset s$ with $\forall s \in \mathbb{S}$.
    \item \textbf{State transition}: new state $s' = \text{exec}_a(s) = s \cup e$ with $\forall s, s' \in \mathbb{S}$.
  \end{itemize}

  For a given symbolic goal set $\mathbb{S}_{\text{goal}} \subset \mathbb{S}$,
  these two operations, allow us to instantiate a search process
  using any tree search algorithm (i.e., depth first, breadth first, etc).
  
  If a skeleton feasible $a_{t:K}^{(R)}$ leading to symbolic goal state $s^g \in \mathbb{S}_{\text{goal}}$ 
  is found, a Non-Linear ``trajectory optimization" Program (NLP) is defined. 
  The NLP considers geometric switches
  in the system kinematics with long-term dependencies.
  In our implementation we use an interior point method
  \cite{Mainprice:20interior, Mainprice:21bewego} to optimize this NLP.

  \subsubsection{Single planning}
  \label{subsubsec:singleplan}

%  Our setup can have multiple tie solution skeletons in the symbolic domain, for example, in a set-table task where a plate and a cup need to be placed, the orders of getting the plate or the cup first are equally valid.
%  However, due to obstacle avoidance and avoiding the human,  multiple alternative solutions may be infeasible at the geometric level. 
%  Hence, our interest is to search for all tie skeletons at symbolic planning, rank the skeletons according to their NLP costs defined in Equation (\ref{eq:geoprob}), and then solve the NLP (Level 3 LGP) instance in the increasingly cost order until a feasible solution is found.
%  The intuition is to choose the least effort skeleton for this daily task setting, which is feasible. 
%  There is a possibility that all skeletons are infeasible, which means the task fails to find a geometrically feasible solution and hence deems unsuccessful.
  
  Given the dynamic LGP instance $I$ and current symbolic state $s_0$, we define the set of all skeletons leading to $\mathbb{S}_{\text{goal}}$ as:
    \begin{multline} \label{eq:skeleton-set}
    \Gamma(s_0, \mathbb{S}_{\text{goal}}, I) = \{a_{1:K}^{(R)}: \forall_{i=1}^K a_i \in A_g, \\
       s_i=\text{exec}_{a_i}(s_{i-1}), s_K \in \mathbb{S}_{\text{goal}}\}
    \end{multline}

    For search efficiency, we define a simple heuristic to guide the search as the symbolic distance from the current state to the goal. The distance is defined as:
    \begin{equation} \label{eq:heuristic}
      h(s) = n(s^g \setminus s)
    \end{equation}
    where $n(s)$ is the cardinality of the state, i.e. the number of grounded propositions.
    Using the heuristic, we search through the symbolic domain for all tie shortest solutions using Dijkstra algorithm. %need citing?. 
   
    Once all tie skeletons are found,
    we rank them by grounding them using simple interpolation paths 
    and computing their costs defined in Equation (\ref{eq:geoprob}).
    We then solve the NLP instance in increasing cost order until a feasible solution is found.
    
     To achieve human avoidance in single planning at the geometric level,
     we populate the human positions as obstacles along the human prediction trajectory. 
    This ensures the worst-case scenario, in which the
    robot finds a collision-free trajectory at the beginning with single planning.  
    
%    To achieve minimal interference with the human in a cooperating task,
%    we infer the current symbolic state from the human prediction trajectory.
%    The mechanism can be task-specific and implied by the domain design.
%    For example, in the set-table task, the proposition describes the object
%    carried by the human in the prediction, grounded from the (human-carry, ?x - object) predicate,
%    which is defined to be in the goal set.
%    In other words, the goal states, which contain (human-carry, cup) or (on, cup, table),
%    are in $\mathbb{S}_{\text{goal}}$.
%    This assumes that the human intentions are always to cooperate to complete the task. 
%    By this mechanism, the robot knows which objects still need to be
%    picked and placed in the task and executes its portion of the task without conflicting with the human.

\begin{algorithm}[t] 
  \SetAlgoLined
  \SetKwInOut{Input}{input}
  \Input{Init state $x_0$, goal set $\mathbb{S}_{\text{goal}}$}
    Deduce symbolic state $s_0$ from $x_0$\;
    Search $\Gamma_0(s_0, \mathbb{S}_{\text{goal}}, I)$\;
    Set $\kappa = a_{1:K_0}^{(R)} \in \Gamma_0$ as best feasible skeleton\;
    Set elapsed time $\tau = 0$\;
    \While{$\mathbb{S}_{\text{goal}}$ not reached at current $t$}{
      Update system kinematics and human position\;
      Deduce current symbolic state $s_t$ from $x_t$\;
      \If{$\mathbf{F}(\kappa, x_t, s_t) = 0$}{
        Search $\Gamma_t(s_t, \mathbb{S}_{\text{goal}}, I)$\;
        Update $\kappa = a_{1:K_t}^{(R)} \in \Gamma_t$\;
        Set elapsed time $\tau = 0$\;
      }
      Optimize NLP (Level 3 in~\cite{Toussaint:15}) of $\kappa$ from time $\tau$\;
      Execute current action of the skeleton $\kappa$\;
      $\tau = \tau + 1$\;
      Wait for next trigger\;
    }
  \caption{Dynamic LGP}
  \label{alg:lgp}
\end{algorithm}

\subsubsection{Dynamic planning}
\label{subsubsec:dynamicplan}

As the actual human behavior may deviate from the prediction,
the motion trajectory or the skeleton $a_{1:K}^{(R)}$ may become sub-optimal or even unfeasible.

Algorithm \ref{alg:lgp} describes the main execution of our Dynamic LGP formulation.
The main idea is to enable the replanning capability for both: symbolic and geometric levels of LGP.

Initially, similar to single planning in Section \ref{subsubsec:singleplan}, the algorithm finds the best (i.e. lowest cost) feasible skeleton at the beginning and sets it to be the current executing skeleton $\kappa \in \Gamma$.

For each replanning trigger, the algorithm validates for actual symbolic and geometric feasibility of the current executing skeleton.
To validate the current executing skeleton, Dynamic LGP first deduces the current symbolic state $s_t$ from the current geometric state $x_t$.
Recall that the geometric state $x_t = (q_t, h_t, o_t)$, concatenates $q_t$ is the robot, $h_t$ the human, and $o_t$ the movable object configurations.
The current human state $h_t$ is retrieved from the human trajectory $\{h_t\}_{t=1}^T$ computed by Hierarchical Motion Prediction module (see Figure~\ref{fig:model}). 
Given the current human state $h_t$, it also updates the collision avoidance inequality constraint $g_{p}$ at the current phase $k(t)$ in the NLP.

As an example of Symbolic Deduction, the predicate (on X Y) is deduced by checking in the system kinematic tree if there is a stable 3D $xy\phi$ joint from X to Y.
Table \ref{tab:symbol-infer} describes our setup symbolic inference for the predicates using the system kinematics. 
Specifically, querying (human-carry, ?x - object) or (agent-carry, ?x - object) predicates can be done using (carry X Y) check. 
This is the mechanism to encode the human intention to the planner symbolically.

\begin{table}
  \centering
  \begin{tabular}{c|c}
    (on X Y) & check if exists a stable 3D $xy\phi$ joint from X to Y\\
    \hline
    (at X Y) & check if $\norm{x_X - x_Y}_2 \leq r | r \in \mathbb{R}$ \\
    \hline
    (carry X Y) & check if exists a stable free joint (6D) from X to Y\\
  \end{tabular}
  \caption{Predicate inference}
  \label{tab:symbol-infer}
\end{table}

%Unlike single planning, in dynamic planning, we can populate the current human position as an obstacle and update its position at the next planning trigger.
%This relaxes the number of constraints for NLPs leading to better performance.

Then, the executing skeleton at the current time~$t$ can be checked for feasibility, both symbolically and geometrically.
Formally, given the current symbolic and geometric state~$x_t,~s_t$, the $\kappa$ skeleton feasibility at the current time~$t$ is defined as:

\begin{equation} \label{eq:feasibility}
\mathbf{F}(\kappa, x_t, s_t) = 
\begin{cases}
  1 & x_0 = x_t, s_0 = s_t  \\
  & \exists x: [t, KT] \to \mathcal{C}: (1b)-(1f)\\
  0 & \text{otherwise}
\end{cases}
\end{equation}

If the skeleton $\kappa$ is feasible, the NLP is optimized for $\kappa$ from the elapsed time $\tau$, i.e., $\forall t \in [\tau, KT]$, given the current system kinematics condition.
Otherwise, it discards the current executing skeleton and resets the elapsed time $\tau = 0$. 
The single planning is then triggered to replan a new skeleton solution set $\Gamma_t(s_t, \mathbb{S}_{\text{goal}}, I)$ and update the new best feasible skeleton.
One may notice that the elapsed time $\tau$ for the current executing skeleton is an implementation detail; however, it plays a crucial role in keeping track remaining execution time for the fixed duration phases of LGP.

\subsection{Long-Term Motion Prediction using Hierarchies}
\label{subsec:motion_prediction}

At the top-level, our hierarchical motion prediction uses
Maximum Entropy Inverse Reinforcement Learning (MaxEnt IRL)~\cite{ziebart2008maximum}
and a low-level which performs full-body motion prediction conditioned on the sequence of sub-goals induced by the sequence of high-level actions $a_{1:K}^{(H)}$ given by the high-level policy.

\begin{table}[t]
  \centering
  \begin{tabular}{c|c}
    Start State & (0, 4, 0, 1, 0, 3, 1, 0, 1, 2)\\
    \hline
    Actions & Go to white shelf \\
                & Pick up cup \\
                & Go to table \\
                & Place \\
    \hline
    End State & (1, 3, 0, 1, 0, 3, 1, 0, 1, 0) \\ 
  \end{tabular}
  \caption{Example high-level trajectory}
  \label{tab:traj}
\end{table}

 \begin{figure}[t]
  \centering
  \includegraphics[width=.7\columnwidth]{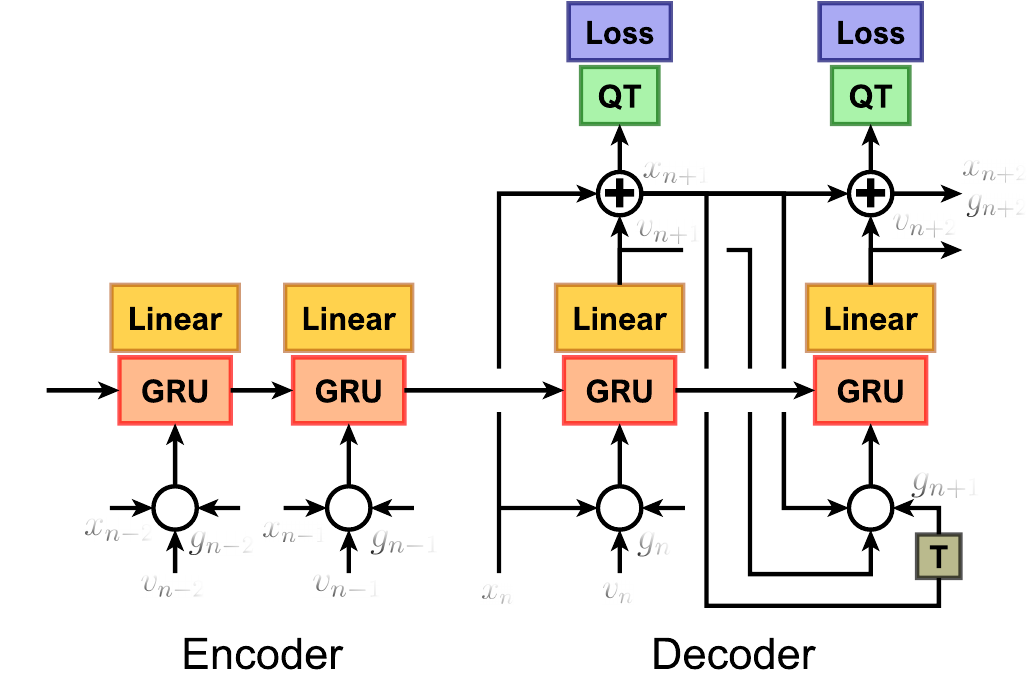}
  \caption{Structure of the VRED~\cite{wang2019vred} with the goal input added.}
  \label{fig:architecture}
\end{figure}

 \subsubsection{Goal-Conditioning}

To be able to use motion prediction as a sub-policy, we do not only
need a sequence-to-sequence mapping but also need it to be goal-conditioned.

Thus, we need a predictive function $h_{t+1:T} = f(h_{0:t}, g^*)$ that computes a trajectory of future human states $h_{t+1:T}$ given previous observed states $h_{0:t}$ and a goal $g^*$.  We use VRED, a recurrent neural network-based model for predicting motion~\cite{wang2019vred} and make it goal-conditioned by adding a three-dimensional position $g_t$ to the input of the network at every timestep (see Figure~\ref{fig:architecture}). The goal input $g_t$ is relative to the coordinate frame of the human and thus changes every timestep.

Particularly, we train two networks, one conditioned on hand target goals for manipulation and another one on pelvis target goals for walking. The network is trained on full-body, kinematic motion trajectories. We use a mean squared distance loss between the base position and a quaternion loss between joint angles. We add an additional loss $l_\text{goal} = | g_T - \phi(x_T)|^2 $ penalizing the distance between the input goal position and the predicted goal position at the last timestep $T$ of the prediction, using a forward kinematic layer~$\phi$.

After training, different trajectories can be generated by varying the goal input manually.
Composing multiple subgoals can be used for sequential long-term motion
predictions of the human.
Note that we could also use other planning-based predictors together with MaxEnt IRL. However, we use VRED due to the scaling property of deep models that learns high-dimensional configuration trajectory of human-motion captures.

\subsubsection{High-level Policy Symbolic State Representation}

We learn a policy $\pi$ that can solve a high-level task. Therefore, we simplify the state-space to a symbolic
representation and use tabular MaxEnt IRL to retrieve our policy.

MaxEnt IRL is based on state frequency calculations. For the MoGaze dataset (see Subsection~\ref{ssec:dataset}),
the discretized state is given by the number of objects on a location and the human position as follows: $s^{(H)} = (cups-table, cups-shelf1, cups-shelf2, plates-table, plates-shelf1, jugbowl-table, jugbowl-shelf1, jugbowl-shelf2, humanPos)$.
The action space is discretized similarly. An example skeleton can be seen in Table~\ref{tab:traj}.

We use heuristics for interfering the exact goal for the human hand or pelvis, for example, by computing the closest point on the table to the human which is not occupied. The heuristics could be further improved by the use of human intention prediction as in~\cite{kratzer2020anticipating}.

The full long-term prediction is achieved by obtaining the skeleton from the high level policy $\pi$,
extracting the goals for the low-level from the heuristics according to the actions in the trajectory,
and using the goal-conditioned RNN to obtain a sequence of
full-body trajectories corresponding to the high-level actions.

\section{Dataset}
\label{ssec:dataset}

We test our framework on the MoGaze dataset~\cite{kratzer2020mogaze}. 
The dataset contains 180 minutes of long, full-body motion sequences for six humans, with 1627 pick and place actions being performed.

Besides human data, the dataset contains object data for two shelves, a table,
and 10 movable objects like cups and plates. 
The participants performed simple manipulation tasks, such as setting up
the table for a fixed number of persons or putting a set of specified objects onto one of the shelves. 
This makes the dataset well suited for our application.

\section{Experiments}
\label{sec:experiments}

Here we report three types of experiments.
First, we report results on hierarchical motion prediction.
Then we report two sets of experiment using LGP, for coordinated
human-robot tasks. In the first, we report a large scale experiment
on 63 segments of setting up the table for 2-3 persons, where we use a
degraded ground-truth trajectory as the prediction.
In the second, we report results on the full pipeline using only 8 segments.

\subsection{Long-Term Motion Prediction using Hierarchies}
%\begin{figure}
%  \centering
%  \begin{subfigure}{.32\columnwidth}
%  \centering
%  \includegraphics[width=\textwidth]{exp1}
%  \end{subfigure}
%  \begin{subfigure}{.32\columnwidth}
%    \centering
%    \includegraphics[width=\textwidth]{exp2}
%  \end{subfigure}
%  \begin{subfigure}{.32\columnwidth}
%    \centering
%    \includegraphics[width=\textwidth]{exp3}
%  \end{subfigure}
%  \caption{Examples of predictions when mapping the learned policy of the IRL algorithm
%to our goal conditioned VRED networks.}
%  \label{fig:res_traj}
%\end{figure}

\begin{figure}[t]
  \centering
%  \includegraphics[width=\columnwidth]{domain.JPG}
%  \Jim{Set the PDDL script as listing}
\begin{lstlisting}
(define (domain set_table)
  (:requirements :typing)
  (:types location object)
  (:constants
    table small_shelf big_shelf - location
    cup_red cup_green cup_blue cup_pink plate_pink plate_red plate_green plate_blue jug bowl - object
  )
  (:predicates
    (agent-at ?l - location)
    (on ?x - object ?l - location)
    (agent-free)
    (agent-avoid-human)
    (agent-carry ?x - object)
    (human-carry ?x - object)
  )

  (:action move
      :parameters (?l - location)
      :precondition ()
      :effect (and (not (agent-at ?*)) (agent-at ?l))
  )
  (:action pick
      :parameters (?x - object ?l - location)
      :precondition (and (agent-at ?l) (on ?x ?l) (not (human-carry ?x)) (at start (agent-free))) 
      :effect (and (not (on ?x ?l)) (not (agent-free)) (agent-carry ?x))
  )
  (:action place
      :parameters (?x - object ?l - location)
      :precondition (and (agent-at ?l) (agent-carry ?x)) 
      :effect (and (not (agent-carry ?x)) (on ?x ?l) (agent-free))
  )
)
\end{lstlisting}
  \caption{PDDL-syntax symbolic domain of set-table task.}
  \label{fig:domain}
\end{figure}

\begin{figure*}
  \begin{tabular}{ccccc}
    \begin{subfigure}{0.18\linewidth}
      \begin{center}
          \includegraphics[width=1.25in, height=1.25in]{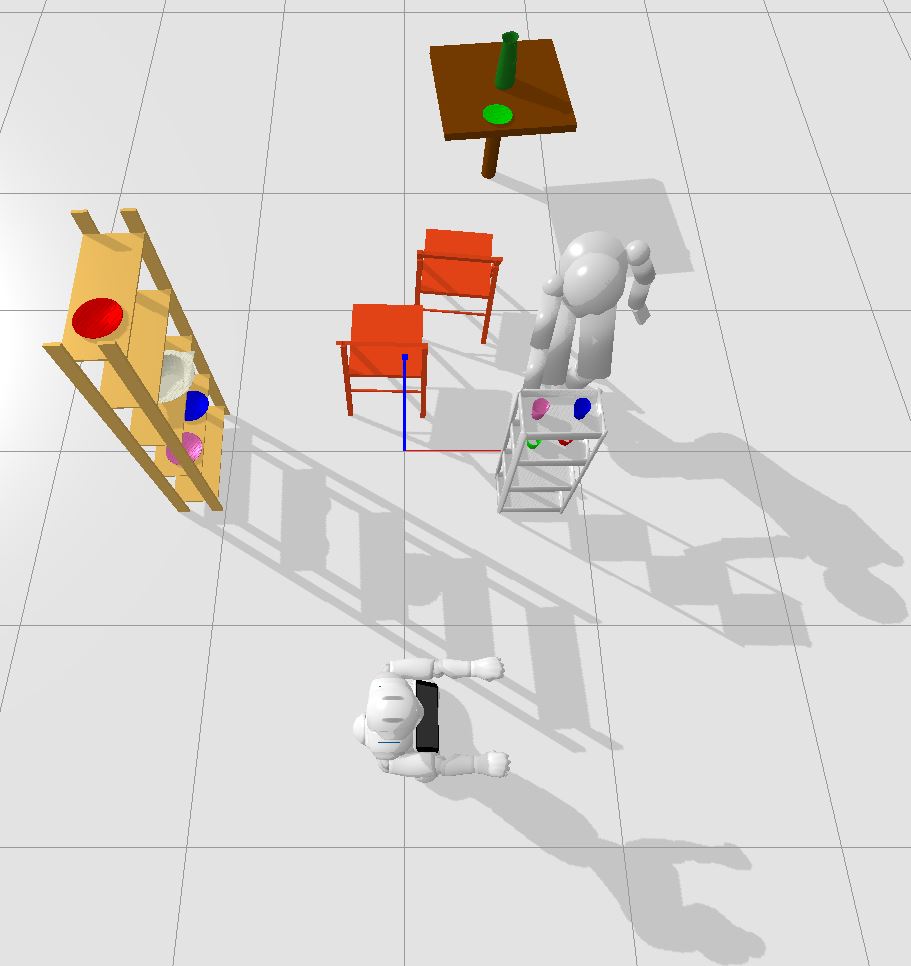}
          \caption{Initial states}
      \end{center}
    \end{subfigure} &
    \begin{subfigure}{0.18\linewidth}
      \begin{center}
          \includegraphics[width=1.25in, height=1.25in]{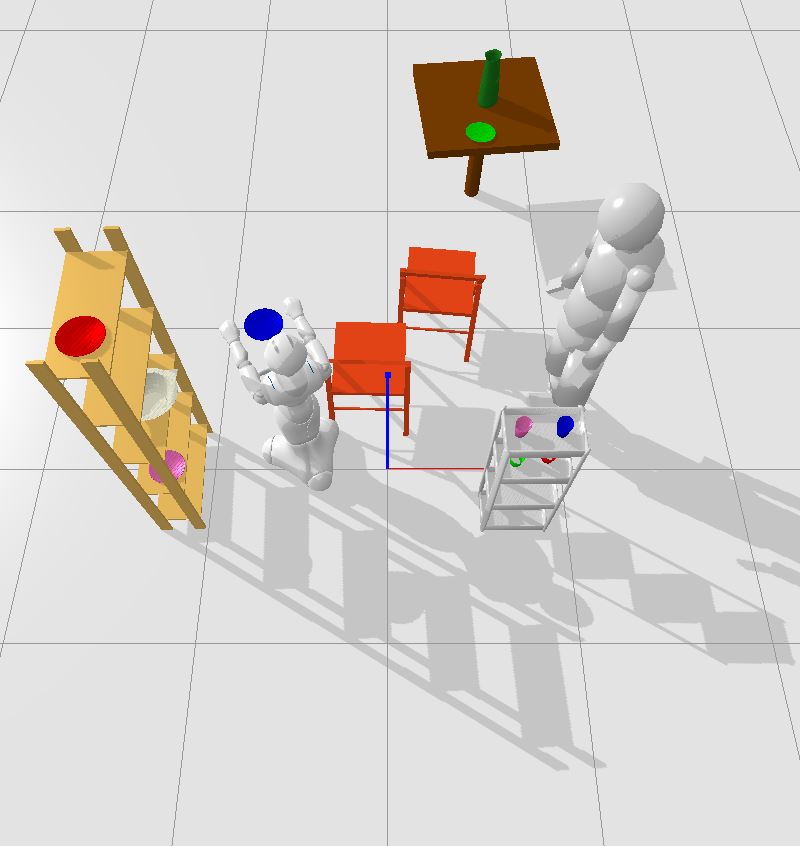}
          \caption{Get blue plate}
      \end{center}
    \end{subfigure} &
    \begin{subfigure}{0.18\linewidth}
      \begin{center}
          \includegraphics[width=1.25in, height=1.25in]{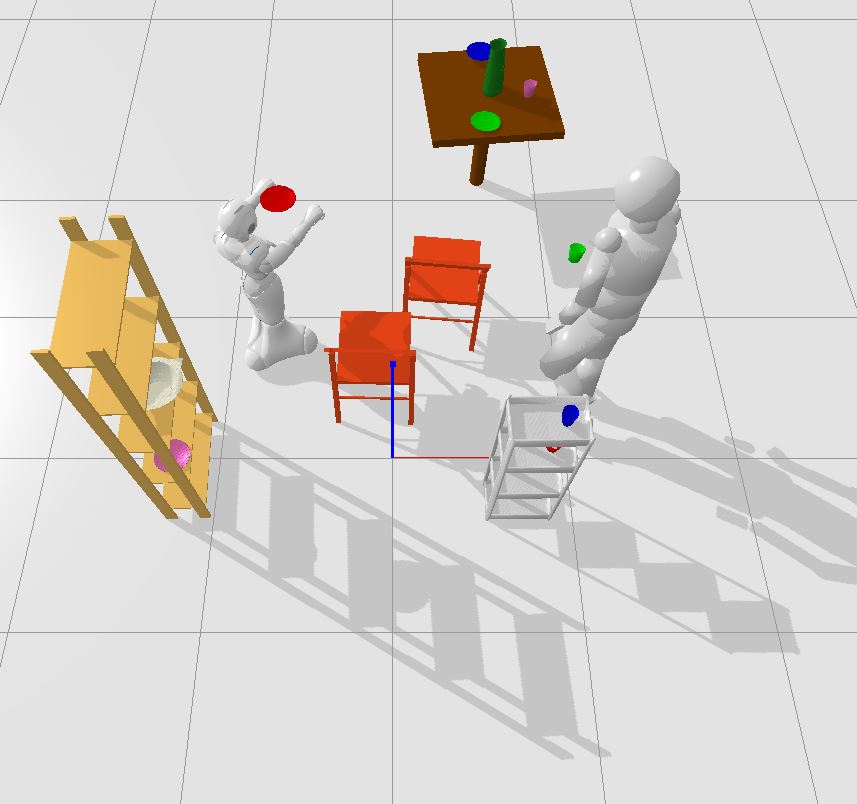}
          \caption{Get red plate}
      \end{center}
    \end{subfigure} &
    \begin{subfigure}{0.18\linewidth}
      \begin{center}
          \includegraphics[width=1.25in, height=1.25in]{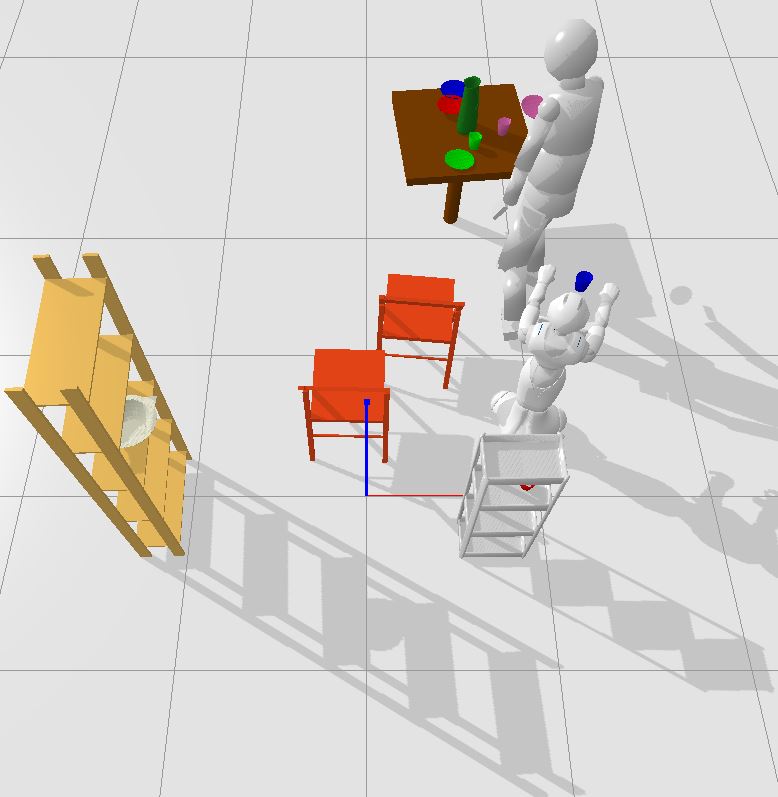}
          \caption{Get blue cup}
      \end{center}
    \end{subfigure} &
    \begin{subfigure}{0.18\linewidth}
      \begin{center}
          \includegraphics[width=1.25in, height=1.25in]{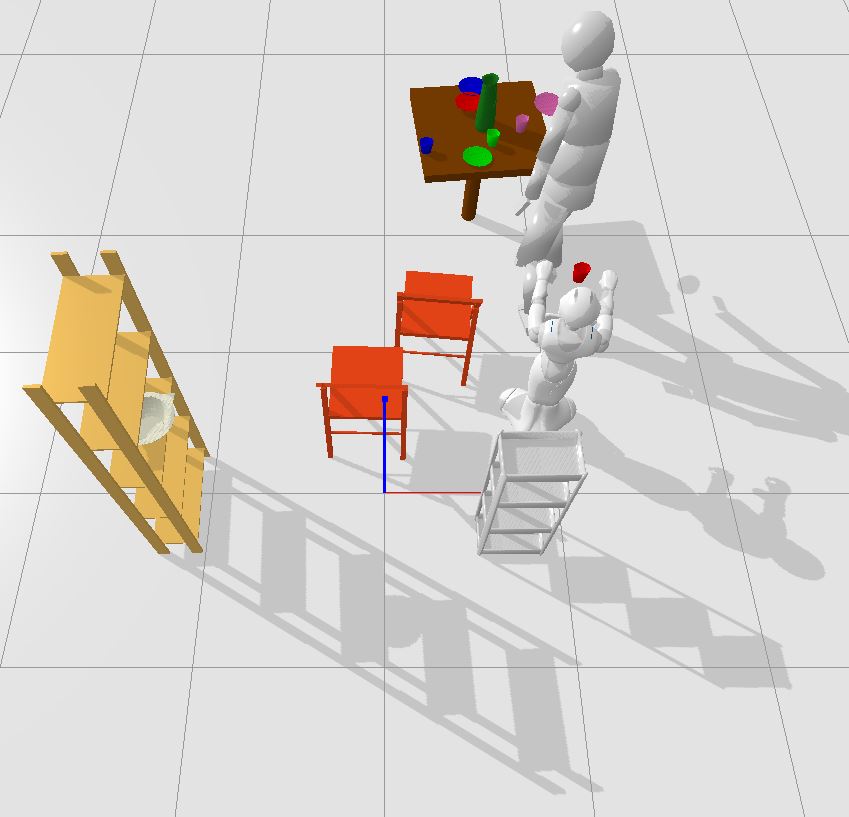}
          \caption{Get red cup}
      \end{center}
    \end{subfigure}\\

  \begin{subfigure}{0.18\linewidth}
    \begin{center}
        \includegraphics[width=1.in, height=1.4in]{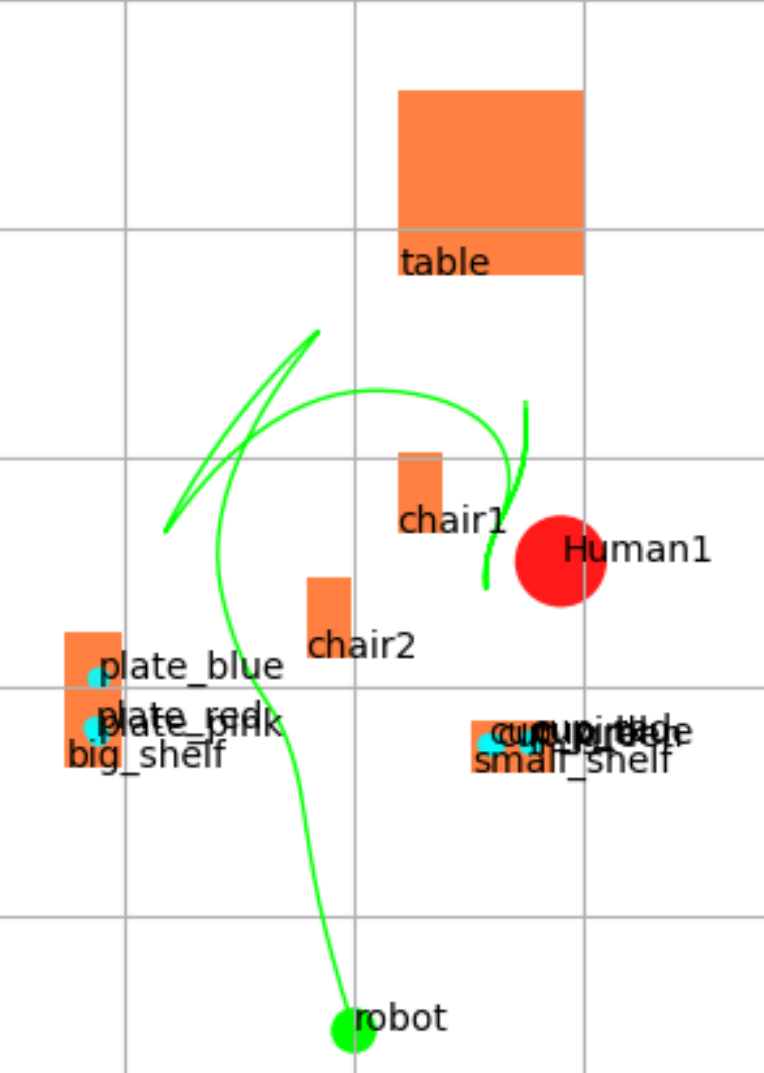}
        \caption{}
    \end{center}
  \end{subfigure} &
  \begin{subfigure}{0.18\linewidth}
    \begin{center}
        \includegraphics[width=1in, height=1.4in]{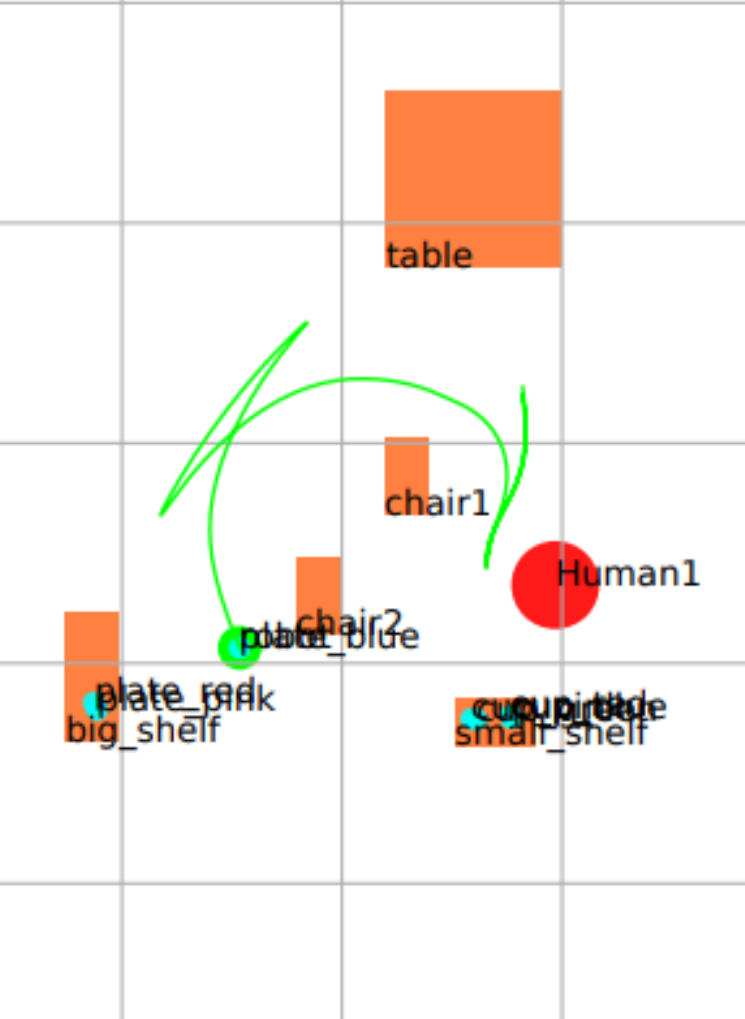}
        \caption{}
    \end{center}
  \end{subfigure} &
  \begin{subfigure}{0.18\linewidth}
    \begin{center}
        \includegraphics[width=1in, height=1.4in]{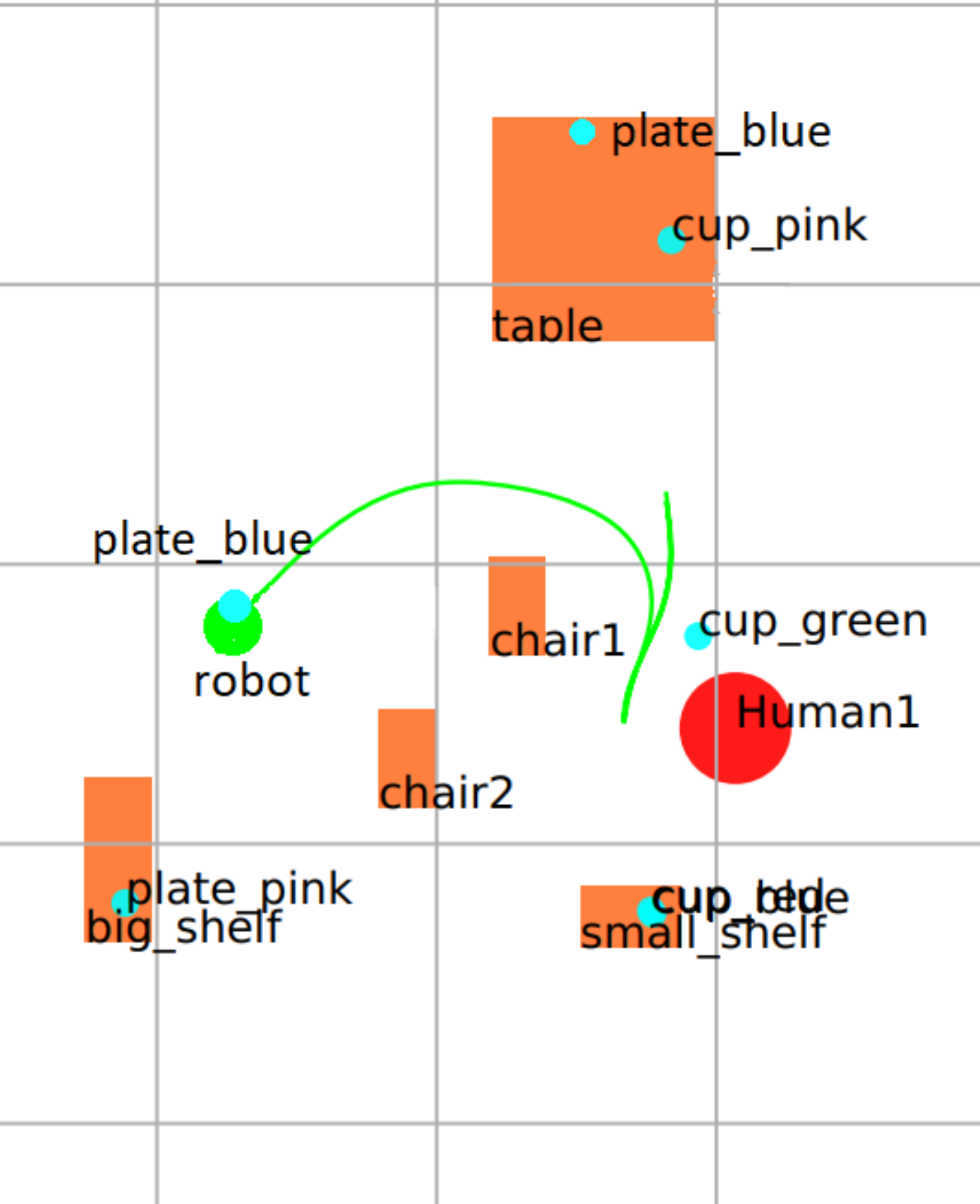}
        \caption{}
    \end{center}
  \end{subfigure} &
  \begin{subfigure}{0.18\linewidth}
    \begin{center}
        \includegraphics[width=1in, height=1.4in]{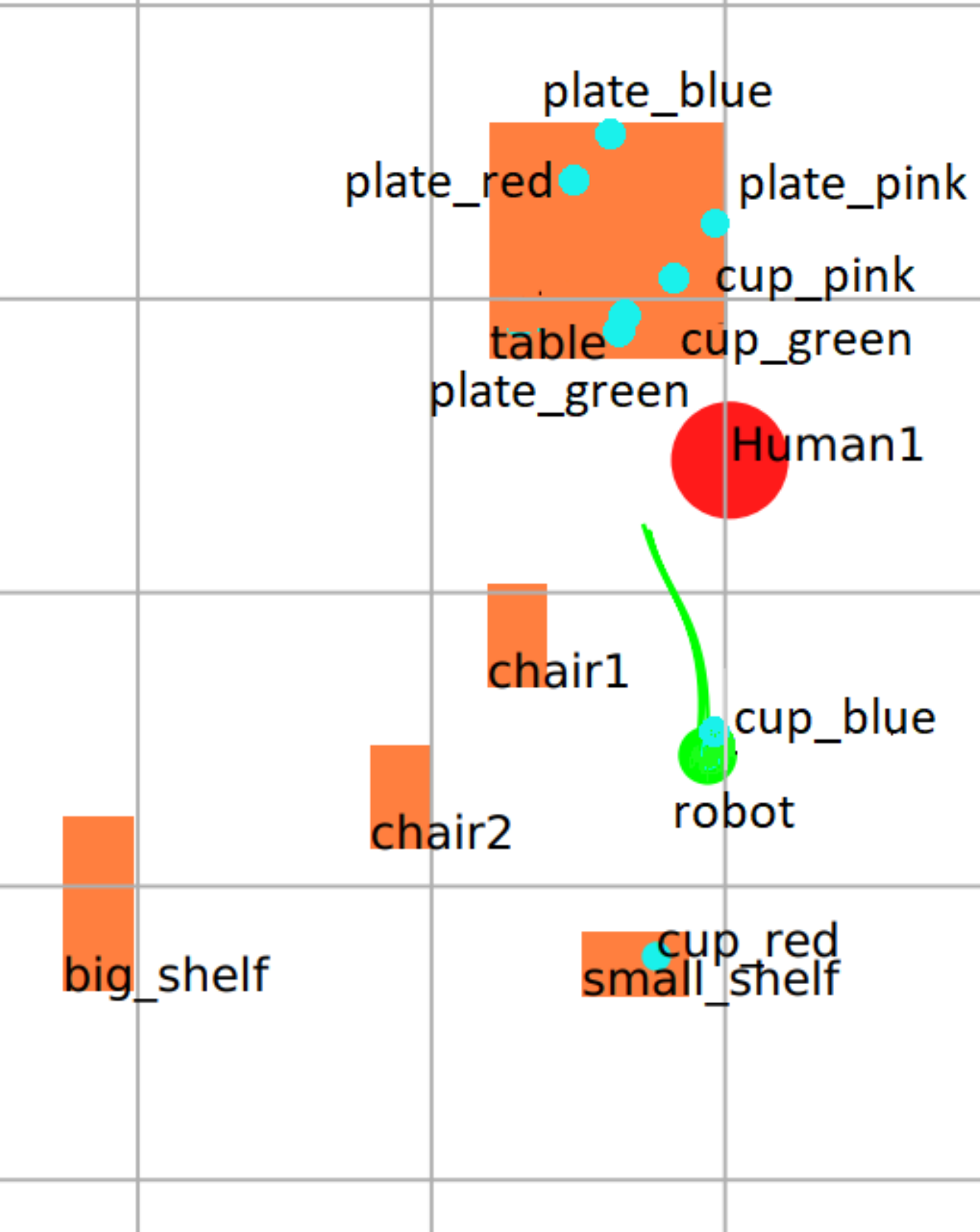}
        \caption{}
    \end{center}
  \end{subfigure} &
  \begin{subfigure}{0.18\linewidth}
    \begin{center}
        \includegraphics[width=1in, height=1.4in]{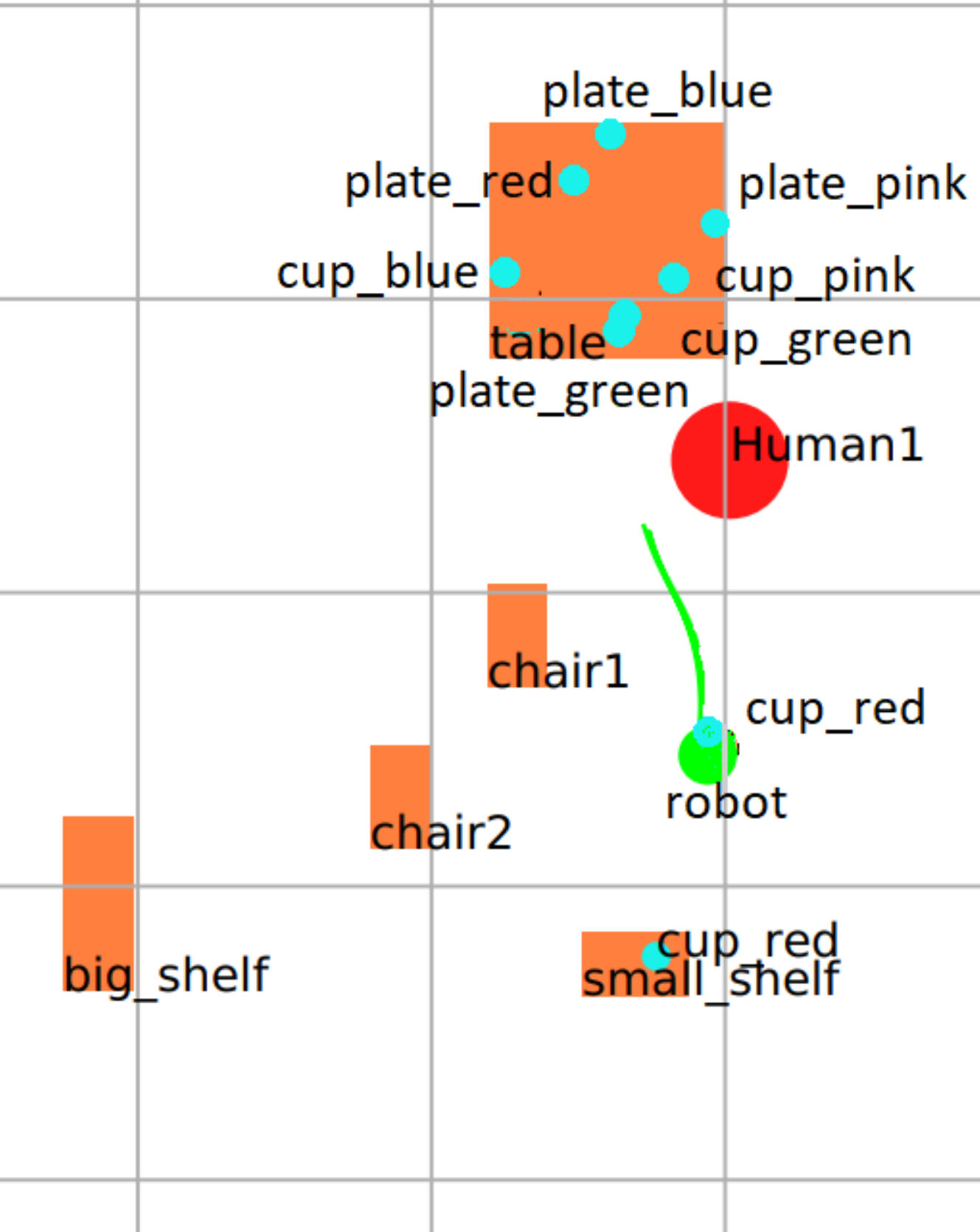}
        \caption{}
    \end{center}
  \end{subfigure}
    
  \end{tabular}
  \caption{Dynamic LGP iterations using human ground truth. Note that in the bottom row images, the robot trajectory is green.}
  \label{fig:lgpdemo}
\end{figure*}

We first compare the original VRED implementation with the VRED conditioned on goal
inputs on the MoGaze dataset.
Results show that the goal-conditioned prediction network achieves both a better joint angular error of $7.99$ instead of $10.14$ and a significantly better base position error of $3.84$ instead of $12.56$, than the network without goal-conditioning.
This is expected because the goal-conditioned network uses oracle information of the goal position.

To test the accuracy of the high-level policy,
 we extracted the task of setting up the table for one person from the dataset.
We then run tabular MaxEnt IRL, showed that the learned policy
solved the task in 80\% of the cases. However, a perfect imitation was achieved solely
in 16\% of the test runs of the cross-validation. This is because the algorithm is limited
by our symbolic state and action representation. Including more complex state features,
e.g., from the 3d environment, could further improve the algorithm.

We also observe that for the walking action the human collides with a chair, as the model is not aware of scene geometry.
In future work this could be avoided by using a prediction method that
accounts for the scene, such as~\cite{kratzer2019prediction}.

\subsection{Dynamic LGP with Human Ground Truth}
\label{ssec:dynamiclgpgt}

To test Dynamic LGP, we design the PDDL-syntax domain following the available objects in the MoGaze dataset. 
Figure \ref{fig:domain} presents the domain for a set-table task,
which consists of a set of necessary predicates and a set of actions,
in which the robot and the human cooperate to pick and place
objects setting the dinner table for 2-3 persons. 

We select 63 dataset segments for this task,
and automate inferring the start symbolic state from the environment kinematics.
We define the robot goal for each segment, e.g. $s_0$ = \{(agent-free), (agent-avoid-human), (on cup-green big-shelf), (on plate-blue small-shelf)\} and the goal state $s_g$ = \{(on cup-green table), (on plate-blue table)\}.

In this experiment we directly feed the human trajectory ground truth into the Dynamic LGP.
We randomly remove a part of the human trajectory in the dataset
for each segment to simulate human prediction data.

The overall task Intersection over Union (IoU)
between the set of objects the human and the robot must place on the table is $0.64 \pm 0.30$. 
In other words, most of the robot task instance has more than half
of the objects to pick and place overlapping with the human task.
Dynamic LGP needs to recognize the overlapping part and plan accordingly. 

The replanning trigger rate is set every 10 timesteps.
Note that if the trigger rate is higher, the executing skeleton would be updated more frequently. Higher rates are desirably to handle more rapid changes
in the environment however it has
to be balanced with the computational capabilities of the robot.
The \textit{move} action has fixed duration of 30 timesteps,
the \textit{pick} and \textit{place} actions both have fixed duration of 5 timesteps.
The sampling rate is 10Hz.

We pause the simulation environment each time the planner is triggered
so that the robot reacts in time to avoid the human obstacle.
An example of a dynamic LGP run is demonstrated in Figure \ref{fig:lgpdemo},
where the task instance consists of 7 objects to pick and place.
The robot executes pick and place for 4 objects, and the human executes the other 3 objects.

\begin{figure}
 \centering
 \begin{subfigure}{.49\columnwidth}
 \centering
 \includegraphics[width=\textwidth]{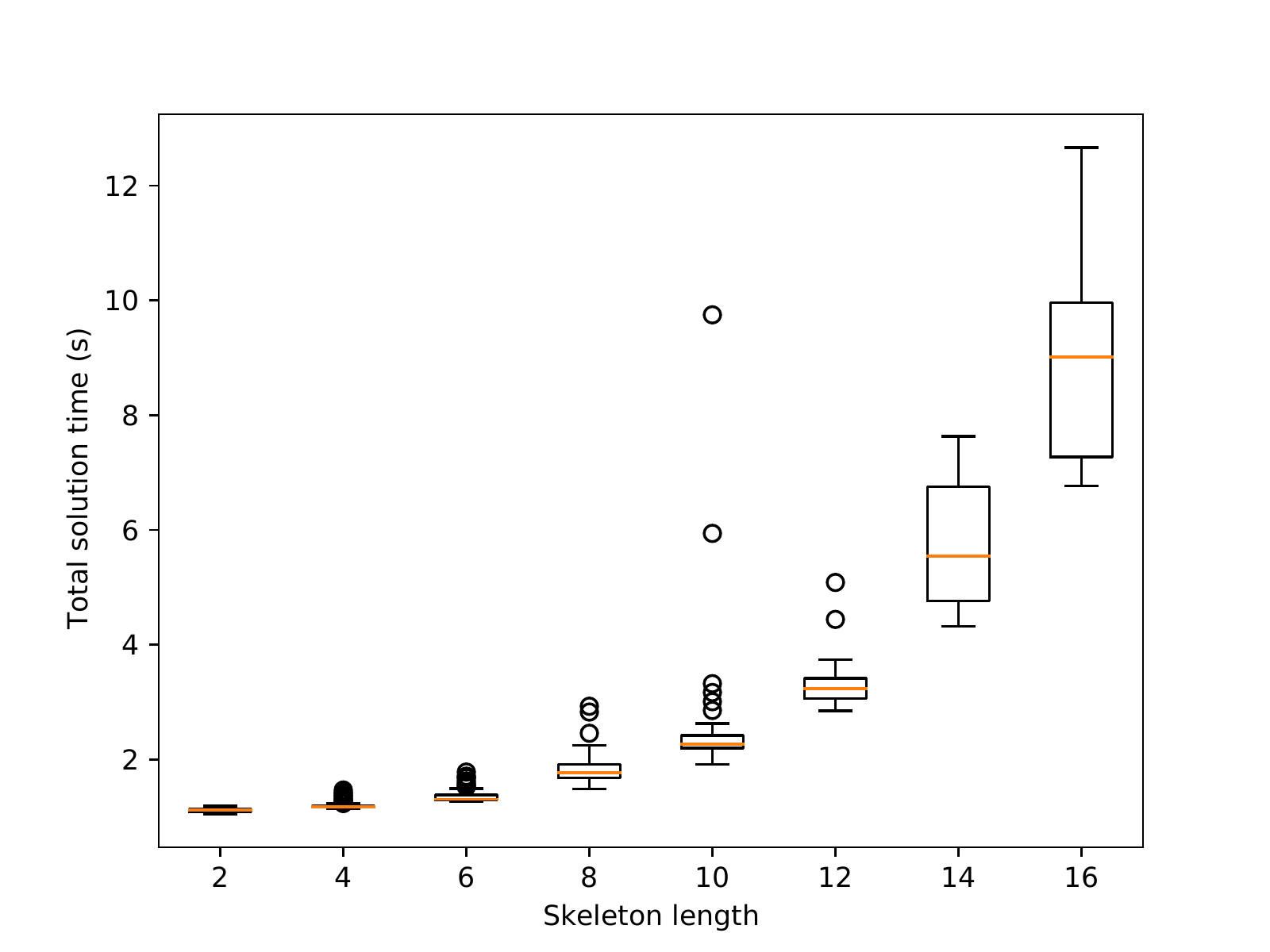}
 \end{subfigure}
 \begin{subfigure}{.49\columnwidth}
   \centering
   \includegraphics[width=\textwidth]{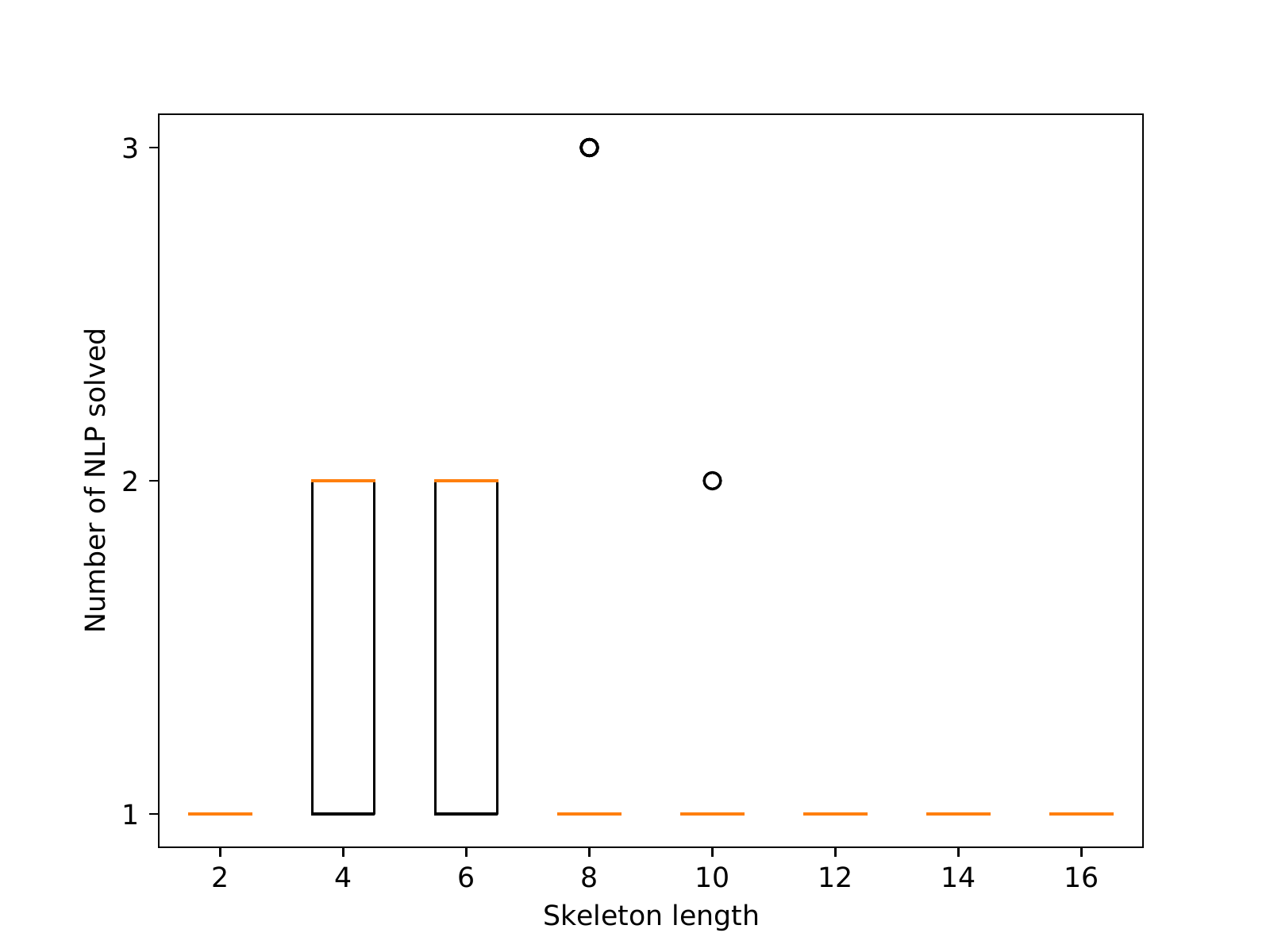}
 \end{subfigure}
 \caption{Total time (left) and number of solved NLPs (right) to find an overall feasible solution over skeleton length.}
 \label{fig:lgp-performace}
 \vspace{-.5cm}
\end{figure}

\begin{figure}
 \centering
 \begin{subfigure}{.49\columnwidth}
 \centering
 \includegraphics[width=\textwidth]{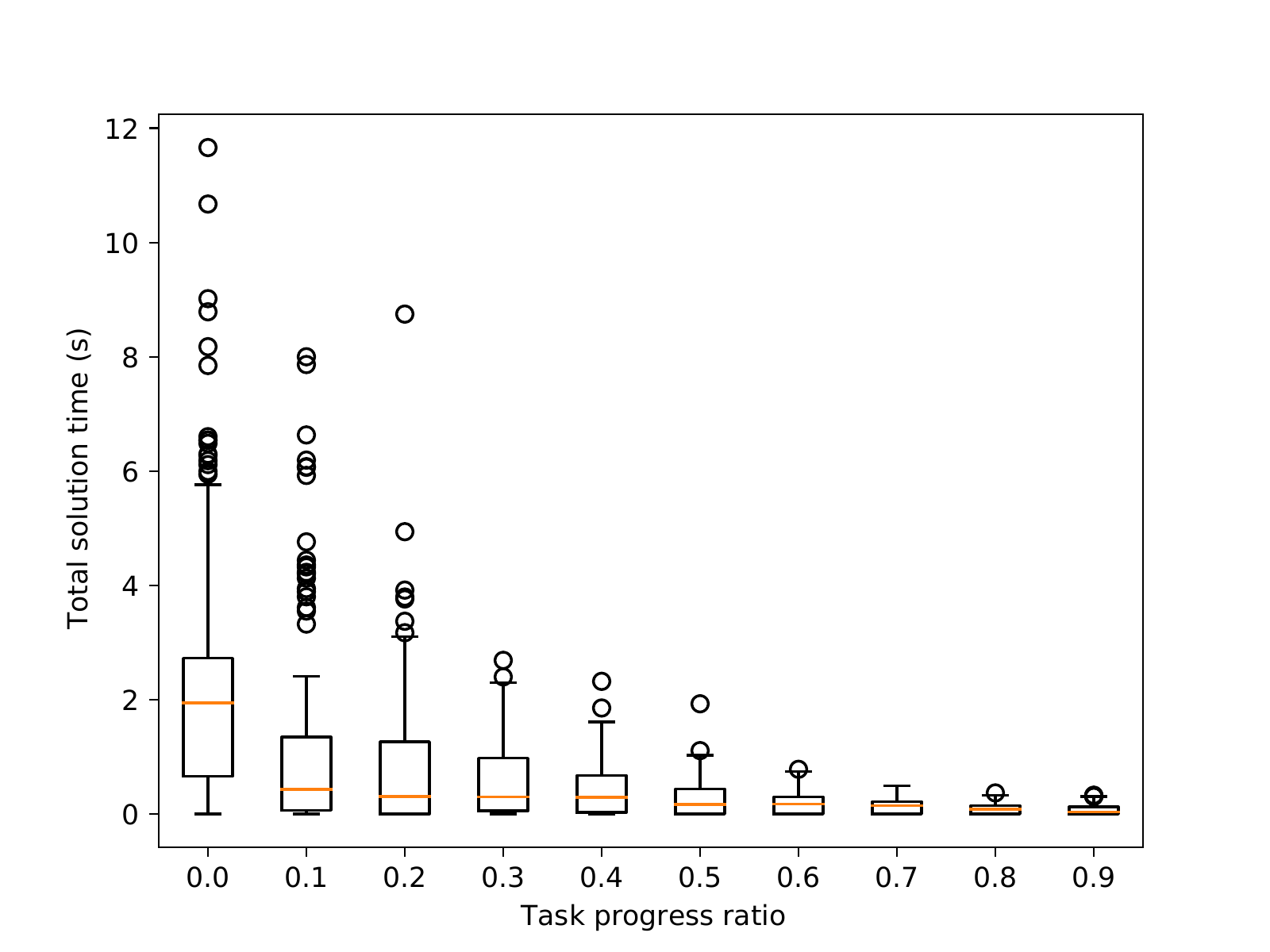}
 \end{subfigure}
 \begin{subfigure}{.49\columnwidth}
   \centering
   \includegraphics[width=\textwidth]{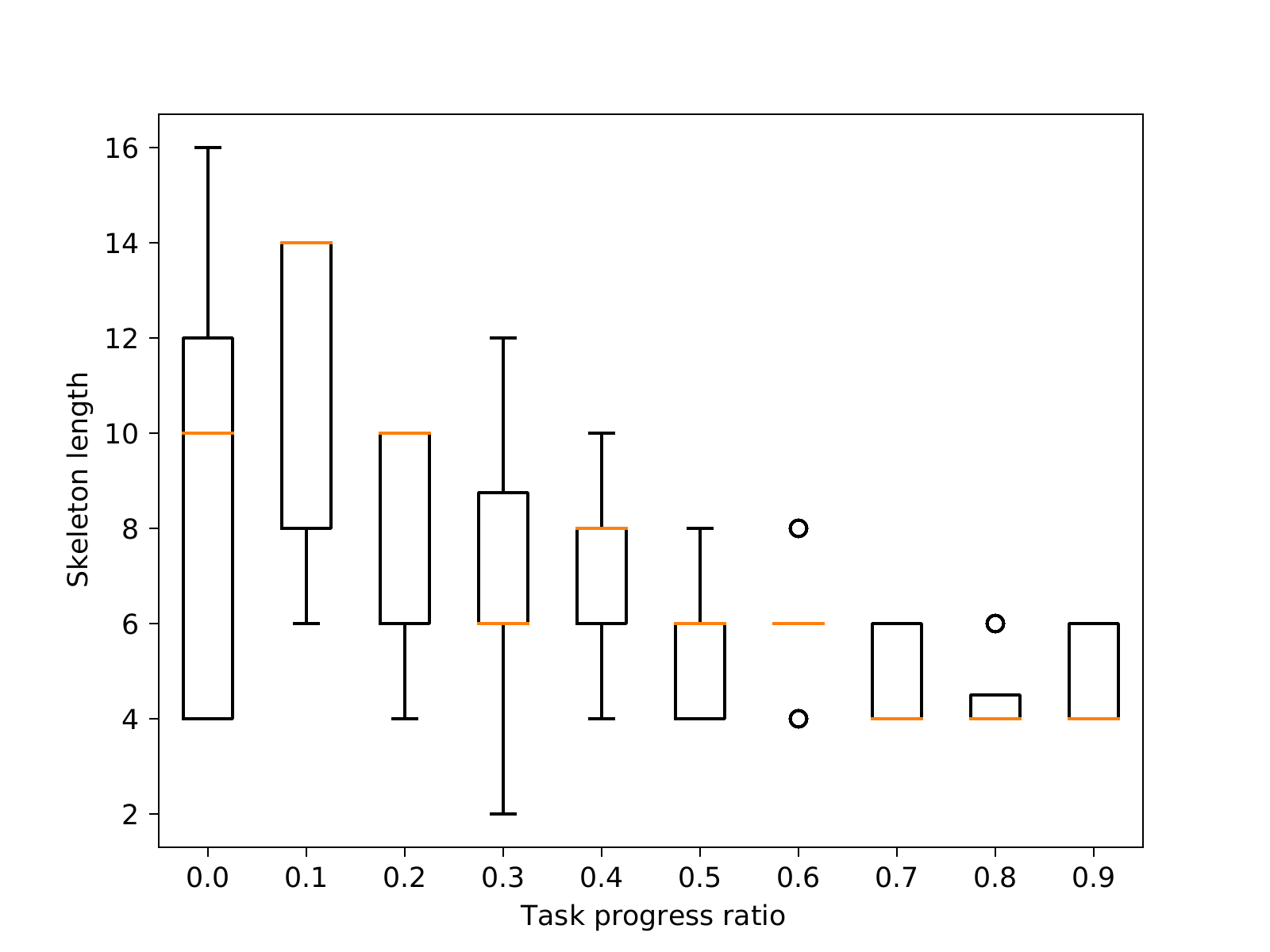}
 \end{subfigure}
 \caption{Total time (left) to find an overall feasible solution and skeleton length (right) over task progress.}
 \label{fig:lgp-over-task}
 \vspace{-.5cm}
\end{figure}

We then run two planning modes; single planning and dynamic planning for each of the 63 segment. 
The task instance is considered successful if,
at the end of the robot trajectory, the deduced symbolic state is in the goal set $\mathbb{S}_{goal}$. 
For dynamic planning, the task fails when the timeout for Algorithm~\ref{alg:lgp}
is reached while the goal set is not satisfied.
For single planning, the task fails when no feasible skeleton is found.

\begin{table}[t!]
  \centering
  \begin{tabular}{c|c|c}
     & Single planning & Dynamic planning\\
    \hline
    Success rate & $84.1\%$ & $95.2\%$\\
    \hline
    Symbolic plan time & $0.032 \pm 0.036(sec)$ & $0.045 \pm 0.053(sec)$\\
    \hline
    Task time reduction & $0.577 \pm 0.107$ & $0.683 \pm 0.099$\\
    \hline
    Path ratio & $1.000$ & $0.584 \pm 0.148$\\
    \hline
    LGP replan count & - & $4.83 \pm 2.21$\\
  \end{tabular}
  \caption{Dynamic LGP with Human Ground Truth}
  \label{tab:gt-statistics}
\end{table}

\begin{table}[t!]
  \centering
  \begin{tabular}{c|c|c}
     & Single planning & Dynamic planning\\
    \hline
    Success rate & $91.2\%$ & $100\%$\\
    \hline
    Symbolic plan time & $0.0005 \pm 0.0001(sec)$ & $0.0006 \pm 0.0002(sec)$\\
    \hline
    Task time reduction & $0.298 \pm 0.078$ & $0.300 \pm 0.100$\\
    \hline
    Path ratio & $1.000$ & $0.626 \pm 0.155$\\
    \hline
    LGP replan count & - & $3.0 \pm 0.87$\\
  \end{tabular}
  \caption{Dynamic LGP with Human Prediction}
  \label{tab:pred-statistics}
\end{table}

Table \ref{tab:gt-statistics} summarizes the statistics in terms
of success rate, symbolic planning time, task time reduction (i.e., the
original time taken by the human to perform the task in the dataset compared
to the execution time with support from the robot) and path ratio (i.e., the ratio of distance
traveled by the robot, with the distance of single planning is the baseline).
Each category is reported in mean and standard deviation over all task instances.
All experiments have been performed with an AMD Ryzen 7 5800X @ 3.8GHz.

As one can see the single planning success rates are
lower than dynamic planning. This 
is expected since the single planning does
not account for potential mismatch between the degraded
ground truth and the actual ground truth.
%It is expected that dynamic planning will have a higher task success ratio of $95.2\%$
%comparing to single planning of $84.1\%$.
%The reason is that single planning considers the worst-case scenario
%where the planner treats all human positions in the trajectory as obstacles;
%this renders many NLPs challenging to solve.

\begin{figure}[t!]
  \centering
  \includegraphics[scale=0.45]{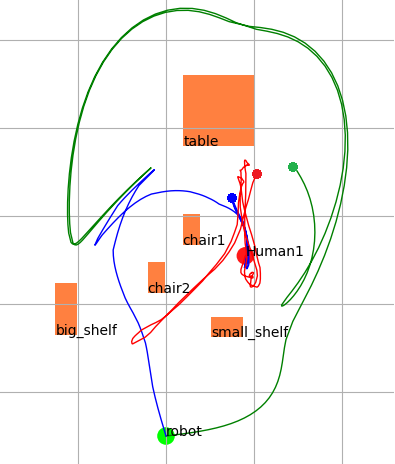}
  \caption{Recorded actual trajectories of single planning (green), dynamic planning (blue) and actual human trajectory (red) on the workspace.}
  \label{fig:trajectories}
\end{figure}

Figure \ref{fig:trajectories} depicts the recorded trajectories of both
planning modes and the human trajectory in a task instance. 
We see that the dynamic mode plans a much shorter trajectory going between
the shelf and the table because the only human obstacle constraint is efficiently updated at every trigger.
We also see in Table~\ref{tab:gt-statistics} that the trajectory length of single planning
is almost twice of dynamic planning (i.e., path ratio).

%However, single planning has a better task complete-time reduction ratio
%between human-robot collaborating scenarios versus only human scenarios. 
%This is because, in dynamic planning, we trigger a symbolic replan
%if the current skeleton is not valid anymore. Thus it takes longer compared to
%single planning to complete the task.

Surprisingly, the experiment shows that the executing symbolic skeleton in
dynamic planning is usually invalid over task progress. Hence full LGP replanning
is triggered frequently.
This shows that the replanning capability is crucial in a dynamic environment,
such as working with humans, since the symbolic state deduced from the environment is rapidly changed.

Figure \ref{fig:lgp-performace} reports both the total solution time and
the number of NLPs that have to be solved to find a feasible solution.
%We refer to total solution time as the total time to compute and rank the NLP
%costs for all possible skeletons plus the time to solve the NLPs until one solution is found.
%Note that the small circles are outliers. 
As can be seen, most of the time only one NLP is needed to reach a feasible solution. 
This implies that usually for this task, the lowest cost skeleton is feasible.
In other cases, more NLPs need to be solved due to the dynamic characteristic of the task.

Generally, the longer the action skeleton, the longer it takes to solve one NLP.
The figure shows a median runtime of about 9 sec for the longest sequence length of 16,
with $\approx$ 450 time steps to optimize in an NLP.
This time step correspond the time discretization of 
interior point trajectory optimization algorithm \cite{Mainprice:20interior, Mainprice:21bewego}.
%However, the longest skeleton only exists at the beginning of the task instance where the
%robot can wait to find an initial solution.
%As in Figure~\ref{fig:lgp-over-task}, as the task progresses, the total solution time to
%find a feasible NLP rapidly decreases to below 1 second since the action skeleton length also decreases.
%This benefits the real-time application, as the planning loop needs to be fast.
%The overall
%framework has reasonable performance to finish the task safely.
Notice in Figure \ref{fig:lgp-over-task} that in some cases,
the action skeleton length increases as the task progresses. 
The skeleton length's median is 6 at task progress ratio 0.3, then increases to 8 at task progress ratio 0.4. 
This implies that the LGP replan sometimes has to resort to longer action skeletons with higher costs as shorter skeletons are infeasible. 

\subsection{Dynamic LGP with Long-Term Prediction}

In this experiment, we choose 8 data segments from MoGaze,
and produce the Long-Term Prediction outputs as described in Section \ref{subsec:motion_prediction}.

For each segment, we run 5 task instances to capture the human motion
prediction statistics due to its stochasticity.
The settings are the same as in Section \ref{ssec:dynamiclgpgt}. 
%However, regarding the human trajectory, we predict a long-term human trajectory at the beginning of the task instance, and then feed the human prediction to the symbolic inference framework of Dynamic LGP.
The overall task IoU between the robot and the human objects is $0.34 \pm 0.13$.
Obviously, this IoU is less than Human Ground Truth experiment since in the Human
Ground Truth experiment the human trajectory is used directly.

Table \ref{tab:pred-statistics} reports task statistics for this experiment. 
%Interestingly, the success rate for dynamic planning is $100\%$ in this case.
%This shows Dynamic LGP is more favorable in dynamic tasks. 
Statistics agree with Table \ref{tab:gt-statistics},
which shows that dynamic planning has higher success rates and produces shorter paths
and needs slightly more time to complete than single planning. 

\begin{figure}
 \centering
 \includegraphics[width=.7\columnwidth]{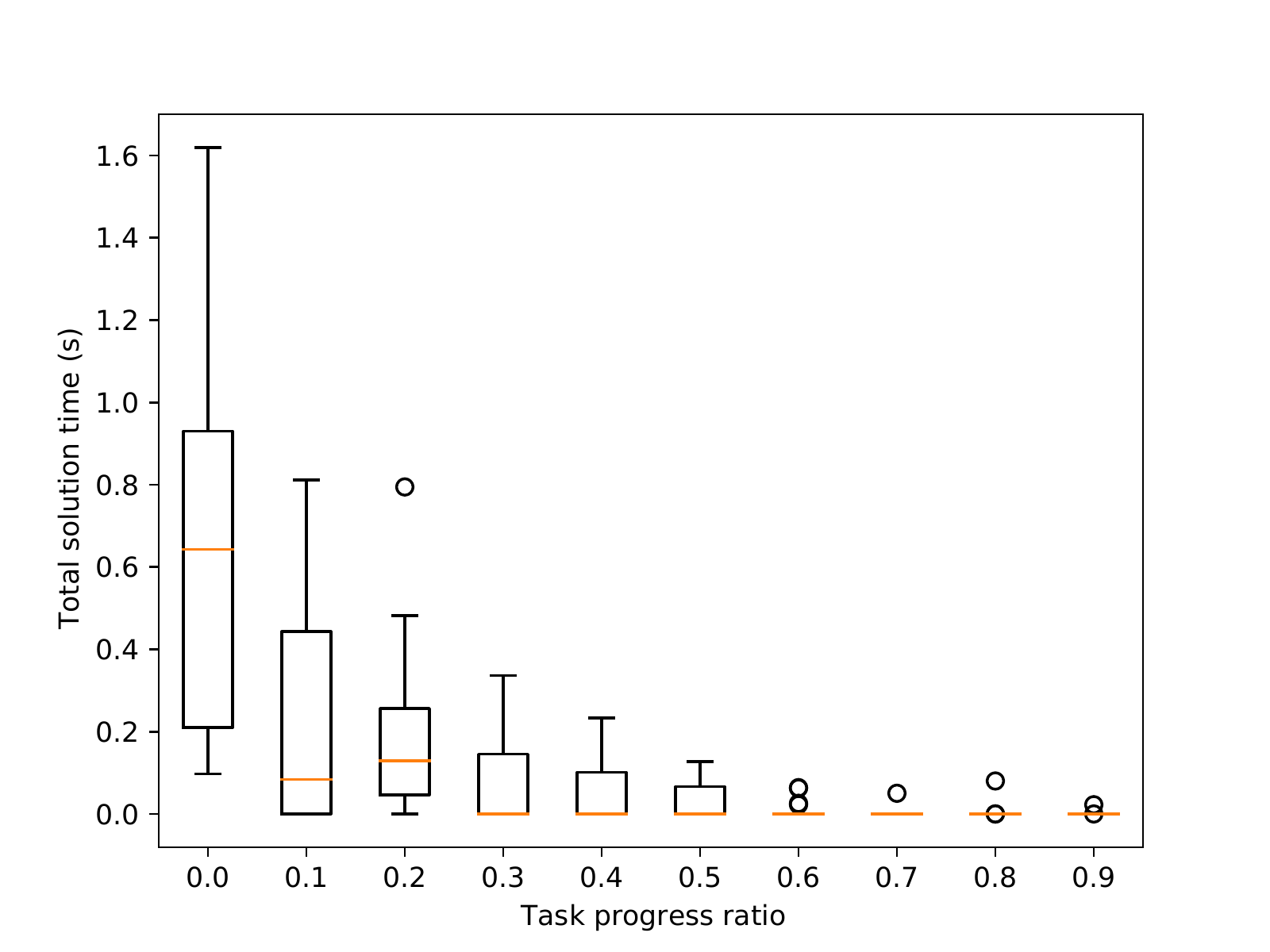}
 \caption{Total time to find an overall feasible solution over task progress with human prediction.}
 \label{fig:pred-over-task}
 \vspace{-.5cm}
\end{figure}

Figure \ref{fig:pred-over-task} also shows decreasing total solution
times as the task progresses with human prediction.
Note that the planning time are an order of magnitude
lower, as the task includes less objects.

%In general, the crucial mechanism is the symbolic inference,
%which informs LGP with the current state of the workspace.
%Through symbolic inference, the human intention encoded in the human trajectory,
%whether ground truth or prediction, is also realized in the task, and thus enables minimal interference.

\section{Conclusions \& Future Work}
\label{sec:conclusions}

Our experiments show that
Dynamic LGP is able to produce plans
that have higher success rate than single planning.
These plans also reduce the total time to execute the task
by a factor approaching 2, which is what one would expected when
two agents collaborate at a task.

The dynamic version of the LGP algorithm we introduce replans periodically to handle the mismatch between the human motion prediction and the actual human motion behavior.
Our experiments demonstrate that the solutions obtained
with human prediction are more efficient and have higher success rate.
We also show that the hierarchical motion prediction approach is capable to produce long-terms human motion predictions.
Futhermore, we design the integration between hierarchical human motion prediction with Dynamic LGP to apply in human-robot collaboration scenarios.

In future work, we aim to include collision avoidance
in our hierarchical motion prediction framework
and produce fullbody robot motions using
a Level 2 NLP for LGP (we refere the reader to \cite{Toussaint:15})
to handle complete robot grasping configurations.
We also plan to port these results to the Pepper robot
of the University of Stuttgart.

%The algorithm is able to plan supporting
%motions that minimally interfere,
%
%In future work, we aim to tackle more tasks
%with more humans involving human-robot coordination.
%We also aim to port this framework to our Pepper robot.
%This is crucial to show the efficacy of the approach
%as there is a behavior feedback loop between the robot
%behavior and the human, which we could not consider
%in this work.

\section*{ACKNOWLEDGMENT}

This work is partially funded by the research alliance ``System Mensch''.
The authors thank the International Max Planck Research School for Intelligent Systems (IMPRS-IS) for supporting Philipp Kratzer.

\balance

\bibliographystyle{IEEEtran}
\bibliography{bibliography}

%\addtolength{\textheight}{-1cm}   % This command serves to balance the column lengths
%                                  % on the last page of the document manually. It shortens
%                                  % the textheight of the last page by a suitable amount.
%                                  % This command does not take effect until the next page
%                                  % so it should come on the page before the last. Make
%                                  % sure that you do not shorten the textheight too much.

\end{document}